\documentclass[10pt,twocolumn,letterpaper]{article}

\usepackage{cvpr}              
\usepackage{graphicx}
\usepackage{booktabs}
\usepackage{multirow}
\usepackage{amsmath}
\usepackage{algorithm}
\usepackage{algorithmic}

\usepackage[dvipsnames]{xcolor}

\definecolor{cvprblue}{rgb}{0.21,0.49,0.74}
\usepackage[pagebackref,breaklinks,colorlinks,citecolor=cvprblue]{hyperref}

\def\paperID{27} 
\def\confName{CVPR}
\def\confYear{2024}

\title{D$^2$-World: An Efficient World Model through Decoupled Dynamic Flow}

\author{
Haiming Zhang\textsuperscript{\rm 1,2},
Xu Yan\textsuperscript{\rm 1}\thanks{Project Lead.},\\
Ying Xue\textsuperscript{\rm 2}, 
Zixuan Guo\textsuperscript{\rm 1}, 
Shuguang Cui\textsuperscript{\rm 2},
Zhen Li\textsuperscript{\rm 2},
Bingbing Liu\textsuperscript{\rm 1}\\
\textsuperscript{\rm 1}Huawei Noah's Ark Lab, 
\textsuperscript{\rm 2}The Chinese University of Hong Kong, Shenzhen\\ 
{\tt\small haimingzhang@link.cuhk.edu.cn}
}

\begin{document}
\maketitle
\begin{abstract}

This technical report summarizes the second-place solution for the Predictive World Model Challenge held at the CVPR-2024 Workshop on Foundation Models for Autonomous Systems.
We introduce \textbf{D$^2$-World}, a novel \textbf{World} model that effectively forecasts future point clouds through \textbf{D}ecoupled \textbf{D}ynamic flow. 
Specifically, the past semantic occupancies are obtained via existing occupancy networks (e.g., BEVDet).
Following this, the occupancy results serve as the input for a single-stage world model, generating future occupancy in a non-autoregressive manner.
To further simplify the task, dynamic voxel decoupling is performed in the world model. 
The model generates future dynamic voxels by warping the existing observations through voxel flow, while remained static voxels can be easily obtained through pose transformation.
As a result, our approach achieves state-of-the-art performance on the OpenScene Predictive World Model benchmark, securing \textbf{second place}, and trains more than \textbf{300\% faster} than the baseline model. Code is available at {\textcolor{magenta}{\url{https://github.com/zhanghm1995/D2-World}}}.
\end{abstract}    
\section{Introduction}
\label{sec:intro}

The predictive world model aims to forecast future states using past observations, playing a crucial role in achieving end-to-end driving systems.
In the CVPR 2024 Predictive World Model Challenge, participants are required to use past image inputs to predict the point cloud of future frames.
This challenge presents two main difficulties: The first is how to effectively train on large-scale data. Given that the OpenScene dataset~\cite{openscene2023} contains 0.6 million frames, the designed model must be efficient. The second challenge is how to predict faithful point clouds through sore visual inputs.
To address these issues, we designed a novel solution that extends beyond the baseline model. 
Regarding the \textbf{Problem I}, we found that the official baseline model (\ie, ViDAR~\cite{yang2023visual}) requires very long training times because it uses all historical frames to predict all future frames in an autoregressive manner. To address this, we designed a solution that divides the entire training process into two parts. The first part trains an occupancy prediction model for single-frame prediction, while the second part uses past occupancy data to predict future point clouds.
Specifically, in the first stage, we utilize an existing occupancy network, such as BEVDet~\cite{huang2021bevdet}, which predicts semantic occupancy by encoding both occupancy states and semantic information within a 3D volume.
In the second stage, a generative world model takes the past occupancy results as input and generates the future occupancy states, which are then rendered into point clouds via differentiable volume rendering.
Through this training paradigm, we increased the training speed by 200\%.

Given the significant development of occupancy networks in the autonomous driving community recently~\cite{huang2021bevdet,li2022bevformer,zhang2024radocc}, for the aforementioned \textbf{Problem II}, we focus on how to construct a world model that maps past occupancy results to future ones.
Our framework leverages the advantages and potential of single-stage video prediction~\cite{ning2022mimo}, enabling the prediction of multiple future volumes in a non-autoregressive manner.
Moreover, we found that directly predicting the occupancy of each frame results in unsatisfactory performance due to the majority of the voxels being empty.
To address this issue, we use the semantic information predicted by the occupancy network to decouple voxels into dynamic and static categories. The world model then only predicts the voxel flow of dynamic objects and warps these voxels accordingly. For static objects, since their global positions remain unchanged, we can easily obtain them through pose transformation.
By leveraging the above components, D$^2$-World surpasses the baseline model by a large margin, achieving a chamfer distance of 0.79 with a single model and securing 2nd place in this challenge.

\begin{figure*}[tbp]
    \begin{center}
        \includegraphics[width=0.95\linewidth]{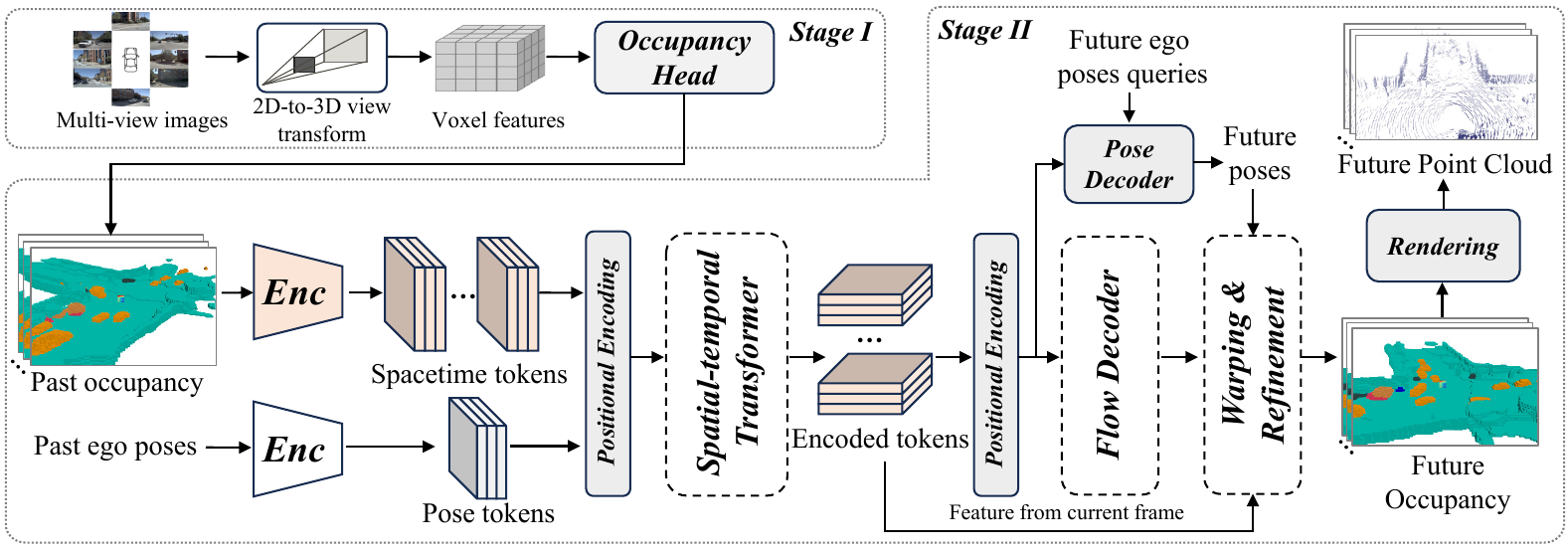}
    \end{center}
    \vspace{-0.6cm}
    \caption{\textbf{The overall pipeline of D$^2$-World.}  In the first stage, we train a single-frame occupancy network, and in the second stage, we train a world model that takes past occupancy as input, forecasting future point clouds.}
    \label{fig:pipeline}
\end{figure*}


\section{Proposed Method}
\label{sec:method}


Our method comprises two stages, and the overall architecture is depicted in Fig.~\ref{fig:pipeline}. Given historical 
$N$ camera images with $T$ timestamps, the first stage predicts occupancy frame-by-frame, aiming to recover a rich 3D dense representation from the 2D images.
In the second stage, we approach this as a 4D point cloud forecasting task. Instead of forecasting the future point cloud in an inefficient autoregressive manner like ViDAR~\cite{yang2023visual}, we design a novel and versatile 4D point cloud forecasting framework that operates in a non-autoregressive manner with decoupled dynamic flow.

\subsection{Stage I: Vision-based Occupancy Prediction}

In this section, we introduce the architecture of the occupancy network, which takes visual images as input and predicts the occupancy state and semantics for a single frame.

\noindent \textbf{Image Encoder}. The image encoder is designed to encode the input multi-camera 2D images into high-level features. Our image encoder comprises a backbone for high-level feature extraction and a neck for multi-resolution feature aggregation. By default, we use the classical ImageNet pre-trained ResNet-50 as the backbone in ablation studies, and Swin-Transformer-B~\cite{liu2021swin} as the backbone for submission. Although employing a stronger image backbone can enhance prediction performance, we considered the trade-offs between resource usage and training time, and ultimately decided against using huge backbones such as InternImage-XL~\cite{wang2022internimage}.

\noindent \textbf{View Transformation.} 
We utilize LSS~\cite{huang2021bevdet} for view transformation, which densely predicts the depth of each pixel through a classification method, allowing us to project the image features into 3D space. Moreover, to introduce the temporal information in our model, we adopt the technique proposed in \cite{li2022bevstereo}, dynamically warping and fusing one historical volume feature to produce a fused feature.

\noindent \textbf{Ocupancy Head.} We adopt the semantic scene completion module proposed in \cite{yan2021sparse} as our occupancy head, which contains several 3D convolutional blocks to learn a local geometric representation. The features from different blocks are concatenated to aggregate information. Finally, a linear projection is utilized to map the features into $C_0$ dimensions, where $C_0$ is the number of classes.

\noindent \textbf{Losses.} 
To alleviate the class-imbalance issue in occupancy prediction, we utilize class-weighted cross-entropy and Lovasz losses. Our multi-task training losses are a combination of occupancy prediction loss and depth loss.

\subsection{Stage II: 4D Occupancy Forecasting}
In this section, we introduce the process of future point cloud forecasting. The framework consists of an occupancy encoder, a flow decoder, flow guided warping and refine, and a rendering process. 

Initially, the 3D occupancy data is preprocessed into spacetime tokens. The spatial-temporal transformer effectively captures the spatial structures and local spatiotemporal dependencies within these tokens.
Following the encoding of historical tokens, the flow decoder is employed to predict future flow in each voxel grid. Then, warping and refinement generate the final occupancy density. 
To fully leverage the temporal information across the entire sequence, we utilize a non-autoregressive approach for decoding, which achieves impressive forecasting performance alongside high efficiency.
Finally, a differentiable volume rendering process is used to generate the point cloud from the predicted occupancy.

\noindent\textbf{3D Occupancy Encoding.}
Given a sequence of historically observed $N_h$ frames 3D occupancy $\mathcal{O}_{T-N_h:T}$, where each occupancy $\mathcal{O}_i \in \mathbb{R}^{H_0 \times W_0 \times D_0}$, we first encode the occupancy sequence into spacetime tokens. 
Here, $H_0$, $W_0$ and $D_0$ represent the resolution of the surrounding space centered on the ego car. Each voxel is assigned as one of $C_0$ classes, denoting whether it is occupied and which semantic category it is occupied with.

To reduce the computational burden, we transform the 3D occupancy in the BEV representation. 
Take a single-frame occupancy as an example, it first uses a learnable class embedding to map the 3D occupancy into occupancy embedding $\hat{\mathbf{y}} \in \mathbb{R}^{H_0 \times W_0 \times D_0 \times C}$.
Then, it reshapes the 3D occupancy embedding along the height dimension to obtain a BEV representation $\mathbf{\tilde{y}} \in \mathbb{R}^{H_0 \times W_0 \times DC}$. 
The BEV embedding then is decomposed into non-overlapping 2D patches $\mathbf{y}_p \in \mathbb{R}^{H \times W \times C^{\prime}}$, where $H=H_0/P$, $W=W_0/P$, $C^{\prime}=P^2\cdot C_0$, and $P$ is the resolution of each image patch.
After that, a lightweight encoder composed of several 2D convolution layers, \ie, Conv2d-GroupNorm-SiLU, is followed to extract the patch embeddings. After considering the sequence of patch embeddings, we obtain the historical occupancy spacetime tokens $\mathbf{{y}} \in \mathbb{R}^{N_h \times H \times W \times C}$.

\noindent\textbf{Ego Pose Encoding.} We represent the ego pose as relative displacements between adjacent frames in the 2D ground plane. Given the historical ego poses, we employ multiple linear layers followed by a ReLU activation function to obtain the ego tokens$\mathbf{e} \in \mathbb{R}^{N_h \times C}$.

\begin{figure}[t]
    \begin{center}
        \includegraphics[width=1.0\linewidth]{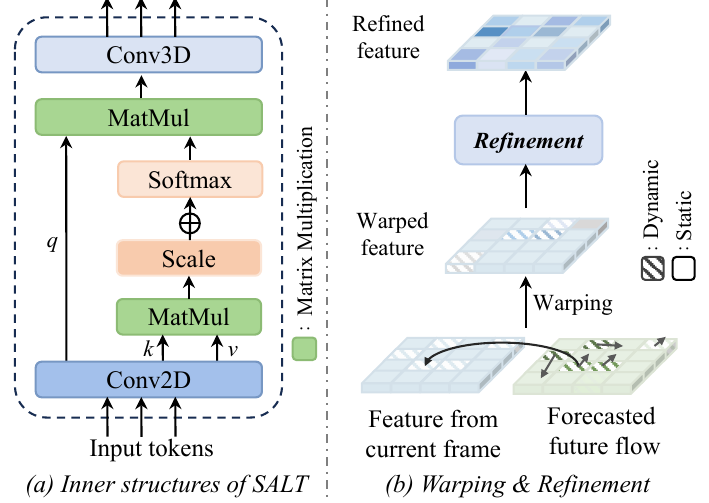}
    \end{center}
    \vspace{-0.3cm}
    \caption{\textbf{Inner structure of SALT \& warping and refinement.} (a) The detailed structures of SALT, which replace the MLP and FFN (Feed Forward Network) in vanilla transformer with 2D convolutions and 3D convolutions respectively for capturing spatial-temporal dependencies. (b) We decouple the flow with the dynamic and static flow and warp the feature of the current frame for forecasting the future frame. The refinement module refines the coarse warping features.}
    \label{fig:flow}
    \vspace{-1em}
\end{figure}

\noindent\textbf{Spatial-Temporal Transformer.}
The spatial-temporal transformer jointly models the evolution of the surrounding scene and plans the future trajectory of the ego vehicle. Inspired by previous works on video prediction~\cite{press2021train, ning2022mimo,fan2022faceformer}, we incorporate several spatial-aware local-temporal (SALT) attention blocks within the Spatial-Temporal Transformer. As shown in Fig.~\ref{fig:flow}(a), in each SALT block, 2D convolution layers are first utilized to generate the query map and paired key-value embeddings for the spacetime tokens, effectively preserving structural information through this spatial-aware CNN operation.
Subsequently, the standard multi-head attention mechanism is employed to capture the temporal correlations between tokens. This approach allows for the learning of temporal correlations while preserving the spatial information of the sequence.
Furthermore, we replace the traditional feed-forward network (FFN) layer with a 3D convolutional neural network (3DCNN) to introduce local temporal clues for enhanced sequential modeling.

\noindent\textbf{Decoupled Dynamic Flow.} 
As illustrated in Fig.~\ref{fig:pipeline} and Fig.~\ref{fig:flow}(b), we design a decoupled dynamic flow to simplify the occupancy forecasting problem.
Specifically, the flow decoder—which comprises multiple stacked SALT blocks—processes the encoded historical BEV features and forecasts the absolute future flows with respect to the current ego coordinate.
Utilizing the occupancy semantics, we decouple the dynamic and static grids, forecasting the future voxel features via the warping operation.
For the dynamic voxels, we transform the absolute flow for each future timestamp using the future ego poses, ensuring alignment with the current frame.
For the static ones, we directly transform them through future ego poses.
Finally, we apply a refinement module composed of several simple CNNs to enhance the coarse warped features.

\noindent\textbf{Rendering \& Losses.}
We utilize the same rendering process and losses as ViDAR~\cite{yang2023visual} for optimizing the point cloud forecasting, which is a ray-wise cross-entropy loss to maximize the response of points along its corresponding ray. For pose regression, we use L1 loss during the training.


\begin{table*}[t]
\small
\centering
\vspace{-0.4cm}
\scriptsize
\begin{tabular}{@{}l|cc|cccccccccc}
\toprule
\multirow{2}{*}{Method} & \multirow{2}{*}{Training Split} & \multirow{2}{*}{Test Split} & \multicolumn{7}{c}{Chamfer Distance ($m^2$) $\downarrow$} \\
& & & 0.5s & 1.0s & 1.5s & 2.0s  & 2.5s & 3.0s & Avg \\
\hline
ViDAR~\cite{yang2023visual} (baseline) & 1/8 Mini & Mini  & 1.34 & 1.43 & 1.51 & 1.60 & 1.71 & 1.86 & 1.58\\
D$^2$-World vanilla (ours) & 1/8 Mini & Mini  & 0.51 & 0.83 & 0.87 & 0.94 & 1.01 & 1.10 & 0.89\\
D$^2$-World (ours) & 1/8 Mini & Mini  & \textbf{0.39} & \textbf{0.74} & \textbf{0.73} & \textbf{0.75} & \textbf{0.80} & \textbf{0.87} & \textbf{0.71}\\
\hline

ViDAR~\cite{yang2023visual}  (baseline)& Mini & Online Server  & 1.32 & 1.41 & 1.49 & 1.60 & 1.73 & 1.93 & 1.59\\
D$^2$-World vanilla (ours) & Mini & Online Server  & 1.19 & 1.47 & 1.50 & 1.57 & 1.65 & 1.79 & 1.53\\\hline
D$^2$-World vanilla (ours) & Full & Online Server & {0.57} & {0.93} & {0.91} & {0.91} & {0.92} & {0.97} & {0.87}\\
D$^2$-World (ours) & Full & Online Server & \textbf{0.56} & \textbf{0.69} & \textbf{0.78} & \textbf{0.84} & \textbf{0.89} & \textbf{0.99} & \textbf{0.79}\\
\bottomrule
\end{tabular}

\caption{\textbf{Poin cloud forecasting performance.}  Best results for each setting are highlighted in bold. D$^2$-World vanilla denotes the model without decoupled dynamic flow.}
\vspace{-0.2cm}
\label{table:pc}
\end{table*}

\begin{table}[ht]
\small
\centering
\scriptsize
\begin{tabular}{@{}l|ccc}
\toprule
Method &  {Hours} & GPU Mem.\\
\hline
ViDAR~\cite{yang2023visual}  (total) & 23.50 & 63G\\
ViDAR~\cite{yang2023visual}  (total) $\dagger$  & 18.50 & 38G\\
\hline
D$^2$-World (stage-I)  & 2.00 & 23G\\
D$^2$-World vanilla (stage-II) & 3.10 & 28G\\
D$^2$-World (stage-II) & 5.14 &32G\\
\hline
D$^2$-World (total) & \textbf{7.14} & \textbf{32G}\\
Proportion & \textbf{~30\%} & \textbf{~51\%}\\

\bottomrule
\end{tabular}
\caption{\textbf{Training efficiency comparisons.} All experiments are trained in 8 GPUs with 24 epochs on 1/8 mini training set. $\dagger$ indicates the efficient version of ViDAR with inferior performance.}
\label{table:efficiency}
\end{table}


\section{Experiments}
\subsection{Experimental Setups}
\noindent \textbf{Dataset.}
We conduct our experiments on the OpenScene dataset~\cite{openscene2023}, which is derived from the nuPlan dataset~\cite{nuplan}.
Due to some scenes in OpenScene lacking corresponding occupancy labels, we ignore these scenes during our experiments.
For submission, the challenge utilizes an online server that provides historical images along with normalized ray directions for point forecasting.

\noindent \textbf{Metric.}
For this challenge, model evaluation is conducted using the Chamfer Distance (CD)~\cite{khurana2023point}. The Chamfer Distance quantifies the similarity between predicted and ground-truth point clouds by computing the average nearest-neighbor distance from points in one set to those in the other set, in both directions.

\noindent \textbf{Training Strategies.} 
During the training process, both stages are trained with AdamW optimizer with gradient clipping and a cyclic learning rate policy. The initial learning rates a $2e^{-4}$ and $1e^{-3}$ for stage I and stage II, respectively.
In stage I, we utilize a total batch size of 24, distributed across 24 NVIDIA V100 GPUs. In stage II, the total batch size is reduced to 16, leveraging 16 NVIDIA V100 GPUs.
For the ablation studies, stage II is trained using 8 NVIDIA V100 GPUs with a total batch size of 8.
Both stages are trained for 24 epochs.

\noindent \textbf{Network Details.} For stage I, the input image resolution is $512\times1408$ incorporating common data augmentation techniques such as flipping and rotation, applied to both the images and the 3D space.
The resolution of the generated 3D voxel grid is $200\times200\times16$.
Prior to feeding the predicted occupancy into stage II, we apply grid sampling operations to align the occupancy annotations from the range of [-50m, -50m, -4m, 50m, 50m, 4m] to the LiDAR point cloud range of [-51.2m, -51.2m, -5.0m, 51.2m, 51.2m, 3.0m].

\subsection{Quantitative Results}

\noindent \textbf{Main Results \& Ablation Study.} The main results are presented in Tab.~\ref{table:pc}. 
In addition to showing the overall performance of our model (D$^2$-World), we also demonstrate the performance of our model without decoupled dynamic flow (D$^2$-World vanilla).
Our method demonstrates superior performance across all timestamps when compared to the baseline model, with further performance enhancements observed upon the introduction of the decoupled dynamic flow. Our best submission ranks 2nd on the leaderboard, achieving a Chamfer Distance (CD) of 0.79, with both stages trained on the full dataset.

\noindent \textbf{Training Efficiency.} To further validate the efficiency of our approach, we compare training hours and GPU memory usage across different models, as shown in Tab.~\ref{table:efficiency}. The baseline method, ViDAR, requires up to 63 GB of GPU memory and 23.50 hours for training. Even its efficient version~\cite{yang2023visual}, which does not supervise all future frames, still demands high GPU memory (38 GB) and considerable training time (18.5 hours).
In contrast, although our method necessitates pre-training an occupancy prediction model, our world model can be trained in approximately 3 hours with only 28 GB of GPU memory under the same conditions. Additionally, our model, even with the decoupled dynamic flow, maintains reasonable training hours and GPU memory.
%

\begin{table}[t]
\small
\centering

\scriptsize
\begin{tabular}{@{}l|ccc}
\toprule
Method & mIoU $\uparrow$ & IoU $\uparrow$ & {Chamfer Distance} ($m^2$) $\downarrow$ \\
\hline
ViDAR~\cite{yang2023visual} (baseline) &  - & - & 1.54 \\ 
Version A & - & 38.29 & 1.68 \\
Version B  & - & 38.76 & 1.64 \\ 
Version C &  - & 40.41 & 1.50 \\
Version D & - & 47.68 & 0.89 \\
Version E (use GT) & - & 100.0 & 0.88 \\
\hline
Version F &  17.06 & 40.41 & 1.09 \\  
Version G &  18.48 & 47.68 & 0.71 \\  
Version H (use GT)  &  100.0 & 100.0 & 0.69 \\
\bottomrule
\end{tabular}
\caption{\textbf{Results analysis.} The effects of occupancy prediction performance.}
\vspace{-1em}
\label{table:abl}
\end{table}

\noindent \textbf{The Effects of Occupancy Performance.} 
The results using different occupancy performances are presented in Tab.~\ref{table:abl}, where only 1/8 mini dataset are used to train.
We first train our world model with binary occupancy prediction (empty and occupied) as inputs. The results from Version A to Version E denote the performance of the world model when the occupancy performance changes.
We find that the world model performs better when the occupancy performance is improved.

Furthermore, introducing decoupled dynamic flow with semantic occupancy inputs yields additional performance enhancements, as shown in Versions F to H.
Interestingly, the performance does not significantly improve even when ground truth occupancy with 100\% mIoU and IoU is used as input. Our analysis indicates that this is due to the inherently sparse nature of point cloud forecasting, which primarily requires predicting the foremost visible surfaces of objects in the 3D space, whereas IoU evaluation for occupancy encompasses the entire dense space.

\section{Conclusion}
In this report, we present our 2nd solution (D$^2$-World) for the Predictive World Model Challenge held in conjunction with the CVPR 2024 workshop.
By reformulating the visual point cloud forecasting predictive world model into vision-based occupancy prediction and 4D point cloud forecasting via decoupled dynamic flow, our solution demonstrates exemplary forecasting performance and significant potential.

{
    \small
    \bibliographystyle{ieeenat_fullname}
    \bibliography{main}

\begin{thebibliography}{14}
\providecommand{\natexlab}[1]{#1}
\providecommand{\url}[1]{\texttt{#1}}
\expandafter\ifx\csname urlstyle\endcsname\relax
  \providecommand{\doi}[1]{doi: #1}\else
  \providecommand{\doi}{doi: \begingroup \urlstyle{rm}\Url}\fi

\bibitem[Caesar et~al.(2021)Caesar, Kabzan, Tan, Fong, Wolff, Lang, Fletcher, Beijbom, and Omari]{nuplan}
Holger Caesar, Juraj Kabzan, Kok~Seang Tan, Whye~Kit Fong, Eric Wolff, Alex Lang, Luke Fletcher, Oscar Beijbom, and Sammy Omari.
\newblock nuplan: A closed-loop ml-based planning benchmark for autonomous vehicles.
\newblock \emph{arXiv preprint arXiv:2106.11810}, 2021.

\bibitem[Contributors(2023)]{openscene2023}
OpenScene Contributors.
\newblock Openscene: The largest up-to-date 3d occupancy prediction benchmark in autonomous driving.
\newblock \url{https://github.com/OpenDriveLab/OpenScene}, 2023.

\bibitem[Fan et~al.(2022)Fan, Lin, Saito, Wang, and Komura]{fan2022faceformer}
Yingruo Fan, Zhaojiang Lin, Jun Saito, Wenping Wang, and Taku Komura.
\newblock Faceformer: Speech-driven 3d facial animation with transformers.
\newblock In \emph{Proceedings of the IEEE/CVF Conference on Computer Vision and Pattern Recognition}, pages 18770--18780, 2022.

\bibitem[Huang et~al.(2021)Huang, Huang, Zhu, Ye, and Du]{huang2021bevdet}
Junjie Huang, Guan Huang, Zheng Zhu, Yun Ye, and Dalong Du.
\newblock Bevdet: High-performance multi-camera 3d object detection in bird-eye-view.
\newblock \emph{arXiv preprint arXiv:2112.11790}, 2021.

\bibitem[Khurana et~al.(2023)Khurana, Hu, Held, and Ramanan]{khurana2023point}
Tarasha Khurana, Peiyun Hu, David Held, and Deva Ramanan.
\newblock {Point Cloud Forecasting as a Proxy for 4D Occupancy Forecasting}.
\newblock In \emph{CVPR}, 2023.

\bibitem[Li et~al.(2022{\natexlab{a}})Li, Bao, Ge, Yang, Sun, and Li]{li2022bevstereo}
Yinhao Li, Han Bao, Zheng Ge, Jinrong Yang, Jianjian Sun, and Zeming Li.
\newblock Bevstereo: Enhancing depth estimation in multi-view 3d object detection with dynamic temporal stereo.
\newblock \emph{arXiv preprint arXiv:2209.10248}, 2022{\natexlab{a}}.

\bibitem[Li et~al.(2022{\natexlab{b}})Li, Wang, Li, Xie, Sima, Lu, Qiao, and Dai]{li2022bevformer}
Zhiqi Li, Wenhai Wang, Hongyang Li, Enze Xie, Chonghao Sima, Tong Lu, Yu Qiao, and Jifeng Dai.
\newblock Bevformer: Learning bird’s-eye-view representation from multi-camera images via spatiotemporal transformers.
\newblock In \emph{European conference on computer vision}, pages 1--18. Springer, 2022{\natexlab{b}}.

\bibitem[Liu et~al.(2021)Liu, Lin, Cao, Hu, Wei, Zhang, Lin, and Guo]{liu2021swin}
Ze Liu, Yutong Lin, Yue Cao, Han Hu, Yixuan Wei, Zheng Zhang, Stephen Lin, and Baining Guo.
\newblock Swin transformer: Hierarchical vision transformer using shifted windows.
\newblock In \emph{Proceedings of the IEEE/CVF international conference on computer vision}, pages 10012--10022, 2021.

\bibitem[Ning et~al.(2022)Ning, Lan, Li, Chen, Chen, Chen, Han, and Cui]{ning2022mimo}
Shuliang Ning, Mengcheng Lan, Yanran Li, Chaofeng Chen, Qian Chen, Xunlai Chen, Xiaoguang Han, and Shuguang Cui.
\newblock Mimo is all you need : A strong multi-in-multi-out baseline for video prediction.
\newblock \emph{arXiv preprint arXiv: 2212.04655}, 2022.

\bibitem[Press et~al.(2021)Press, Smith, and Lewis]{press2021train}
Ofir Press, Noah~A Smith, and Mike Lewis.
\newblock Train short, test long: Attention with linear biases enables input length extrapolation.
\newblock \emph{arXiv preprint arXiv:2108.12409}, 2021.

\bibitem[Wang et~al.(2022)Wang, Dai, Chen, Huang, Li, Zhu, Hu, Lu, Lu, Li, et~al.]{wang2022internimage}
Wenhai Wang, Jifeng Dai, Zhe Chen, Zhenhang Huang, Zhiqi Li, Xizhou Zhu, Xiaowei Hu, Tong Lu, Lewei Lu, Hongsheng Li, et~al.
\newblock Internimage: Exploring large-scale vision foundation models with deformable convolutions.
\newblock \emph{arXiv preprint arXiv:2211.05778}, 2022.

\bibitem[Yan et~al.(2021)Yan, Gao, Li, Zhang, Li, Huang, and Cui]{yan2021sparse}
Xu Yan, Jiantao Gao, Jie Li, Ruimao Zhang, Zhen Li, Rui Huang, and Shuguang Cui.
\newblock Sparse single sweep lidar point cloud segmentation via learning contextual shape priors from scene completion.
\newblock In \emph{Proceedings of the AAAI Conference on Artificial Intelligence}, pages 3101--3109, 2021.

\bibitem[Yang et~al.(2023)Yang, Chen, Sun, and Li]{yang2023visual}
Zetong Yang, Li Chen, Yanan Sun, and Hongyang Li.
\newblock Visual point cloud forecasting enables scalable autonomous driving.
\newblock \emph{arXiv preprint arXiv:2312.17655}, 2023.

\bibitem[Zhang et~al.(2024)Zhang, Yan, Bai, Gao, Wang, Liu, Cui, and Li]{zhang2024radocc}
Haiming Zhang, Xu Yan, Dongfeng Bai, Jiantao Gao, Pan Wang, Bingbing Liu, Shuguang Cui, and Zhen Li.
\newblock Radocc: Learning cross-modality occupancy knowledge through rendering assisted distillation.
\newblock In \emph{Proceedings of the AAAI Conference on Artificial Intelligence}, pages 7060--7068, 2024.

\end{thebibliography}
}


\end{document}